\documentclass{article}
\usepackage{pgfplots}
\pgfplotsset{compat=1.18}
\usepackage{tikz}
\usepackage{PRIMEarxiv}
\usepackage{caption}
\usepackage{subcaption}
\usepackage{tabularx} 
\usepackage{graphicx}
\usepackage{caption}
\usepackage{booktabs}
\usepackage{geometry}
\usepackage{graphicx}
\usepackage{caption}
\usepackage{subcaption}
\usepackage{tabularray}
 \usepackage{natbib}
\usepackage{pgfplots}
\pgfplotsset{compat=1.18}
\usepackage{sansmath} 
\sansmath

\usepackage[utf8]{inputenc} 
\usepackage[T1]{fontenc}    
\usepackage{hyperref}       
\usepackage{url}            
\usepackage{booktabs}       
\usepackage{amsfonts}       
\usepackage{nicefrac}       
\usepackage{microtype}      
\usepackage{lipsum}
\usepackage{fancyhdr}       
\usepackage{graphicx}       
\graphicspath{{media/}}     
\usepackage{amsmath}
\title{Latent Prototype Routing: Achieving Near-Perfect Load Balancing in Mixture-of-Experts
\thanks{Preprint - Work in Progress. Code: \href{https://github.com/RANDO11199/LatentPrototypeRouter/}{Here}}
}
\author{
  Jiajie Yang \\
  \texttt{jiajie.y@wustl.edu} \\
}
\begin{document}
\maketitle
\begin{figure}[ht]
    \centering
    \begin{subfigure}[b]{0.49\linewidth}
        \includegraphics[width=\textwidth]{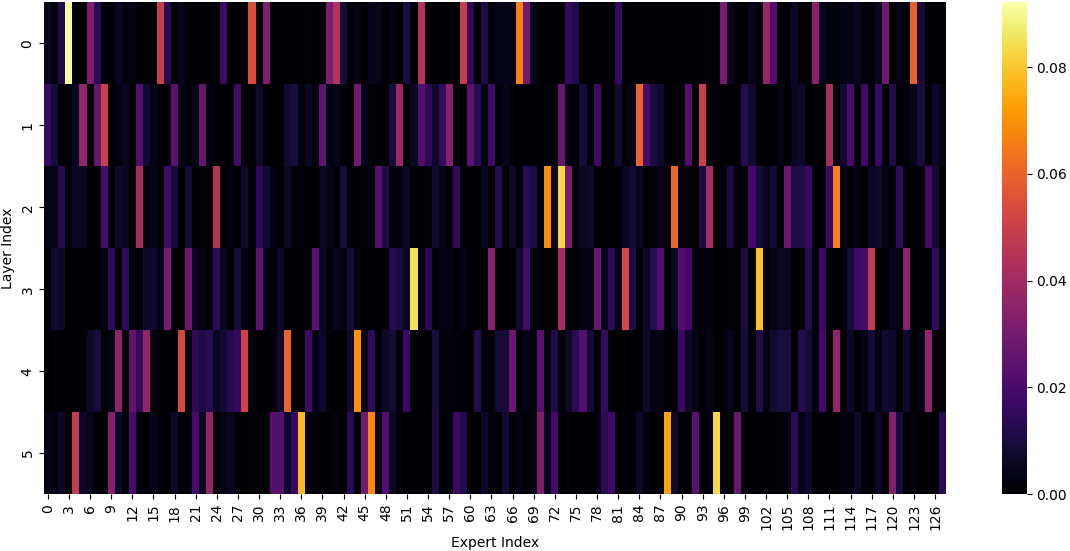}
        \caption{The normalized expert load visualization across different layer of Qwen3Moe.}
        \label{fig:sub1}
    \end{subfigure}
    \hfill
    \begin{subfigure}[b]{0.49\linewidth}
        \includegraphics[width=\textwidth]{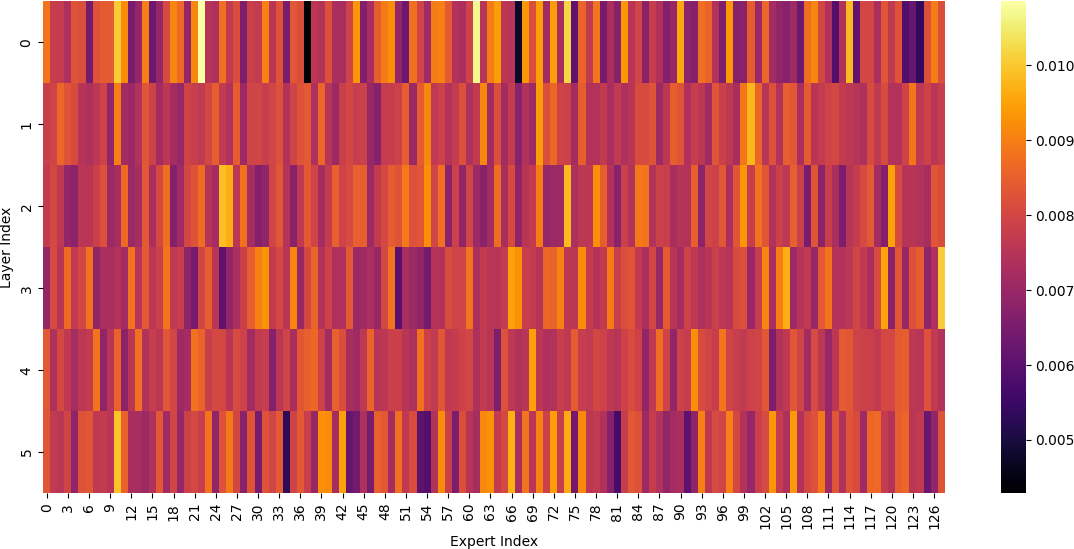}
        \caption{The normalized expert load visualization across different layer of Qwen3Moe-LPR.}
        \label{fig:sub2}
    \end{subfigure}
    \caption{Comparison of loda balance between vanilla router and Latent Prototype Router on the Qwen3Moe model. (a) Few experts are frequently activated compared to others (b) Experts are activated with balanced bahavior}
    \label{fig:test}
\end{figure}

\begin{abstract}

Mixture-of-Experts (MoE) architectures have emerged as a key strategy for scaling large language models (LLMs) efficiently. However, current MoE systems suffer from severe load imbalance, where only a small subset of experts is consistently activated during training and inference, leading to significant underutilization of model capacity and computational resources. In this work, we revisit expert routing through a clustering perspective and propose Latent Prototype Routing (LPR), a novel routing framework that generalizes existing approaches while promoting balanced expert utilization without compromising downstream performance. Extensive experiments across multiple open-source MoE models—including DeepSeek-V3, Qwen3-MoE, and Mixtral—demonstrate that LPR reduces the Gini coefficient of expert load from 0.70 to 0.035 on average, improves the min-max expert load ratio from 1e-6 to 0.70, achieving near-perfect load balancing.
\end{abstract}
\section{Introduction}
Recent advances in Transformer-based architectures \cite{vaswani2017attention} have propelled large language models (LLMs) to unprecedented performance across natural language processing tasks—from machine translation and question answering to code generation. State-of-the-art models like Qwen, Deepseek, and LLaMA now feature hundreds of billions of parameters, enabling sophisticated contextual understanding. However, this scaling comes with a steep computational cost: naive parameter scaling multiplies training and inference overheads, straining the resources of even major research entities.

Mixture-of-Experts (MoE) architectures offer a breakthrough by activating only a subset of sub-networks ("experts") per input token, reducing per-token FLOPs while maintaining model capacity. Early sparsely-gated MoE designs \cite{shazeer2017outrageously} demonstrated that thousands of experts can be trained with marginal compute increases, yielding superior generalization on large-scale benchmarks. Systems like GShard \cite{lepikhin2020gshard}, Switch Transformer \cite{fedus2021switch}, and BASE\cite{base2021} further optimized expert sharding and load-balancing strategies, enabling trillion-parameter models across hundreds of accelerators. Notably, Deepseek recently proposed an auxiliary-free loss to balance loads without degrading performance\cite{wang2024auxiliarylossfreeloadbalancingstrategy}

Despite these advancements, \textbf{expert routing} remains a critical bottleneck. Existing research has predominantly focused on auxiliary losses or learning strategies, while neglecting the routing algorithm itself. This oversight leads to severe load imbalance in industrial MoE models: mainstream implementations exhibit GINI indices of 0.8--0.9 (without auxiliary losses) or 0.7--0.8 (with aux-losses)\cite{wang2024auxiliarylossfreeloadbalancingstrategy}, with min-max load ratios approaching 0. (i.e. min expert loading /max expert loading).  Recent work on differentiable routing—such as DSelect-k \cite{hazimeh2021dselectkdifferentiableselectionmixture} and ReMoE \cite{wang2025remoefullydifferentiablemixtureofexperts}—relaxes hard top-k selection but often relies on global softmax (sacrificing sparsity) or complex auxiliary objectives. 

A high GINI index and low min-max load ratio pose critical challenges. The information absorption capacity of neural networks is fundamentally constrained by both their total parameters \cite{shazeer2017outrageously} and the effective utilization of Mixture-of-Experts (MoE) parameters. A high GINI index indicates uneven parameter activation—prior work has shown that a subset of experts often handle the majority of computations\cite{lepikhin2020gshard}, while a low min-max ratio reveals severe expert underutilization or overutilization. This imbalance creates a dual bottleneck:

\begin{enumerate}

\item \textbf{Knowledge storage bottleneck}: Sparse expert activation limits the model’s ability to encode diverse semantic knowledge, affecting its generalization capacity.

\item \textbf{Hardware-software mismatch}: Heterogeneous load distributions cause GPU memory fragmentation and pipeline stalls, increasing end-to-end latency and introducing severe problem into hardware implementation.
\end{enumerate}

\paragraph{Our Perspective}
We reframe routing as a clustering problem in latent distribution, moving beyond traditional logit-based classification paradigms. Conventional routing algorithms operate through high-dimensional prototype matching (using dot-product similarity between token embeddings and expert keys), which we generalize by introducing:
\begin{enumerate}
    \item \textbf{Nonlinear projection}: A learnable encoder that maps tokens into a low-dimensional semantic latent space
    \item \textbf{Discriminative similarity metrics}: Stable alternatives to dot-product matching (e.g., cosine distance, cross-attention)
    \item \textbf{Distributional regularization}: Explicit constraints on token latent representations and expert prototypes to enforce separable cluster structures.
\end{enumerate}

Formally, we model the token distribution as $\mathcal{P}(z|x,\theta)$ and expert prototypes as $\{\mu_e\}_{e=1}^E$, minimizing the geometric distance or distribution distance between $z$ and $\{\mu_e\}$ to ensure cluster alignment. 

Our principal contributions include:
\begin{enumerate}
    \item \textbf{A unified routing framework} that generalizes existing algorithms while maintaining consistent training-inference behavior
    \item \textbf{Token clustering with distributional constraints}: Novel regularization techniques that enforce separability in latent cluster structures
    \item \textbf{Hybrid Prototype Refinement}: We integrate two complementary optimization strategies:
    \begin{itemize}
        \item \textit{Gradient-based geometric alignment}: Minimizes distributional/geometric distances between tokens $z$ and prototypes $\{\mu_e\}$ via backpropagation
        \item \textit{Non-gradient-based prototype adaptation}: An Exponential Moving Average (EMA) mechanism that incrementally aligns expert prototypes with assigned tokens using the update rule:
        \begin{equation*}
        \mu_e^{(t+1)} \leftarrow \lambda \mu_e^{(t)} + (1-\lambda) \cdot \mathbb{E}_{z \in \mathcal{B}_e}[z]
        \end{equation*}
        where $\mathcal{B}_e$ denotes tokens assigned to expert $e$ for hard version and all tokens for soft version and $\lambda$ is the momentum coefficient. This dual approach simultaneously enhances cluster cohesion while promoting load-balanced expert specialization.  
    \end{itemize} 
    \item \textbf{Comprehensive empirical validation}: Demonstrated reduction of Gini coefficients by 95\% (from 0.7 to 0.035) and improvement of min-max load ratios from near-zero ($10^{-6}$) to 0.7, outperforming state-of-the-art routing schemes across multiple models.
\end{enumerate}
\section{Method}
In this section, we first revisit the standard Transformer and Mixture-of-Experts (MoE) architectures, then analyze current routing methods through the lens of clustering and prototype matching, and finally introduce our \emph{Latent Prototype Routing} (LPR) framework.

\subsection{Preliminaries: Transformer and MoE}

\paragraph{Transformer Encoder Layer.}  
A Transformer layer~\cite{vaswani2017attention} processes input token embeddings $\{\mathbf{h}_t\}_{t=1}^T\subset\mathbb{R}^d$ via a self‐attention and a feed‐forward network (FFN):
\begin{align}
  \mathbf{a}_t &= \mathrm{MultiHeadAttention}(\mathbf{h}_t, \mathbf{H}, \mathbf{H}),\\
  \mathbf{h}'_t &= \mathrm{LayerNorm}\bigl(\mathbf{h}_t + \mathbf{a}_t\bigr),\\
  \mathbf{f}_t &= \mathrm{FFN}(\mathbf{h}'_t)
    = W_2\,\mathrm{ReLU}(W_1 \mathbf{h}'_t + b_1) + b_2,\\
  \mathbf{h}_{t}^{\text{out}}
    &= \mathrm{LayerNorm}\bigl(\mathbf{h}'_t + \mathbf{f}_t\bigr).
\end{align}

\paragraph{Mixture‐of‐Experts (MoE).}  
An MoE layer replaces the standard FFN with $M$ expert networks $\{E_i\}_{i=1}^M$.  A \emph{router} computes a weight vector $\mathbf{w}_t\in\mathbb{R}^M$ for each token:
\begin{align}
  \mathbf{g}_t &= \mathrm{Router}(\mathbf{h}'_t),\\
  w_{t,i} &= 
    \begin{cases}
      \frac{\exp(g_{t,i})}{\sum_{j\in\mathcal{S}_t}\exp(g_{t,j})},
        & i\in\mathcal{S}_t,\\
      0,&\text{otherwise},
    \end{cases}
\end{align}
where $\mathcal{S}_t\subset\{1,\dots,M\}$ is the set of Top-$k$ experts for token~$t$.  The MoE output is then
\begin{equation}
  \mathbf{y}_t \;=\; \sum_{i\in\mathcal{S}_t} w_{t,i}\,E_i(\mathbf{h}'_t).
\end{equation}
Standard routers use a linear projection followed by Top-$k$ softmax, but suffer from non‐differentiability and load imbalance.

\subsection{Routing as Clustering}

Routing in Mixture-of-Experts (MoE) can naturally be viewed as a clustering problem. Let $\mathbf{x}\in\mathbb{R}^d$ denote the token representation and $\mathbf{w}$ the expert keys representing cluster centroids. Given a token representation $\mathbf{x} \in \mathbb{R}^d$, and a set of expert keys $\{\mathbf{w}_i\}_{i=1}^{M}$, the router computes similarity scores between $\mathbf{x}$ and each prototype, typically via inner product:
\[
s_i = \langle \mathbf{x}, \mathbf{w}_i \rangle, \quad \text{for } i = 1, \ldots, M.
\]
Based on these scores, the top-$k$ experts are selected for token processing. This is mathematically assigning $\mathbf{x}$ to the centroids with the most similar direction and magnitude. This clustering view enables explicit modeling of token distributions and incorporation of realistic prior assumptions into the routing mechanism.
\subsubsection{Token Distribution Observations and Routing Implications}
We make two key observations about token distributions in large language models:
\begin{itemize}
\item \textbf{Clusterability:} Token representations naturally form a limited number of semantically coherent clusters in the embedding space. From this perspective, the number of experts can be viewed as our model’s hypothesis on the number of token clusters.
\item \textbf{Imbalanced Frequencies:} The sizes of these clusters are highly skewed; different types of tokens occur with vastly different frequencies. This imbalance induces an inherent tension between load balancing and expert specialization.
\end{itemize}

These observations reflect the underlying linguistic structure, where common syntactic or semantic patterns occur frequently while rare or highly specialized concepts appear infrequently.
Crucially, they also imply a fundamental trade-off between \emph{expert specialization} and \emph{load balancing}. Routing tokens purely by semantic similarity yields experts that specialize on coherent subsets of tokens, but leads to highly uneven loads. Conversely, enforcing strict uniformity across experts improves balance at the expense of semantic purity and diversity.

We argue that expert specialization is as important as balanced load for downstream performance. Hence, we advocate a data-driven, clustering-based routing mechanism that incorporates explicit regularization to encourage balanced expert utilization without sacrificing specialization. During training, as token embeddings and expert keys co-adapt, the router is able to discover a semantically meaningful partition of the representation space that respects both model capacity and the natural heterogeneity of the data.

\subsection{Vanilla Routing Algorithm}
Standard MoE routers typically implement the following pipeline. Given a batch $X \in \mathbb{R}^{B\times d}$ and expert keys $W\in\mathbb{R}^{M\times d}$, we compute
\[
S = XW^\top,
\quad
P = \mathrm{softmax}(S)
\]
where $P$ denotes the normalized assignment scores. In this setup, each row of $W$ acts as an expert key, and the dot-product matches each token against all experts. Although straightforward, this \emph{vanilla} routing strategy suffers from several well-known limitations:
\begin{itemize}
\item \textbf{High-variance scores in high-dimensional spaces.} When $d\gg 1$, the variance of dot-products scales as
\begin{equation}
\text{Var}(XW^\top),\propto, d \cdot \sigma_{x}^2 \sigma_{w}^2,
\end{equation}
making the signal-to-noise ratio deteriorate and causing the softmax output to collapse toward an argmax-like behavior. Even small noise overwhelms semantic signals.
\item \textbf{Prototype collapse.} Without explicit regularization, optimization often encourages all expert keys ${\mathbf{w}_i}$ to align along a dominant subspace that captures the most frequent token patterns, reducing diversity across experts and undermines the benefits of specialization.\cite{beyer1999nearest}
\item \textbf{Curse of dimensionality.} Similarity measures become less discriminative as $d$ grows \cite{aggarwal2001surprising}. Distance concentration in high-dimensional spaces reduces the capacity of dot-product or cosine similarity to separate different token types.
\item \textbf{Linear encoding bottleneck.} Vanilla routers often rely on identity or single-layer projections to produce token embeddings. This constrains the representation capacity and does little to encourage token separation or reduce geometric collapse.
\item \textbf{Lack of explicit clustering constraints.} Token–expert mappings are learned without explicit modeling or regulations of token or expert distributions. This can lead to a mismatch between token semantics and expert assignments, with token embeddings becoming skewed, collapsed, thereby undermining expert specialization and model capacity.
\end{itemize}

\subsection{Latent Prototype Router}\label{sec:vdtr}
\begin{figure}[t]
    \centering
    \includegraphics[width=0.7\textwidth]{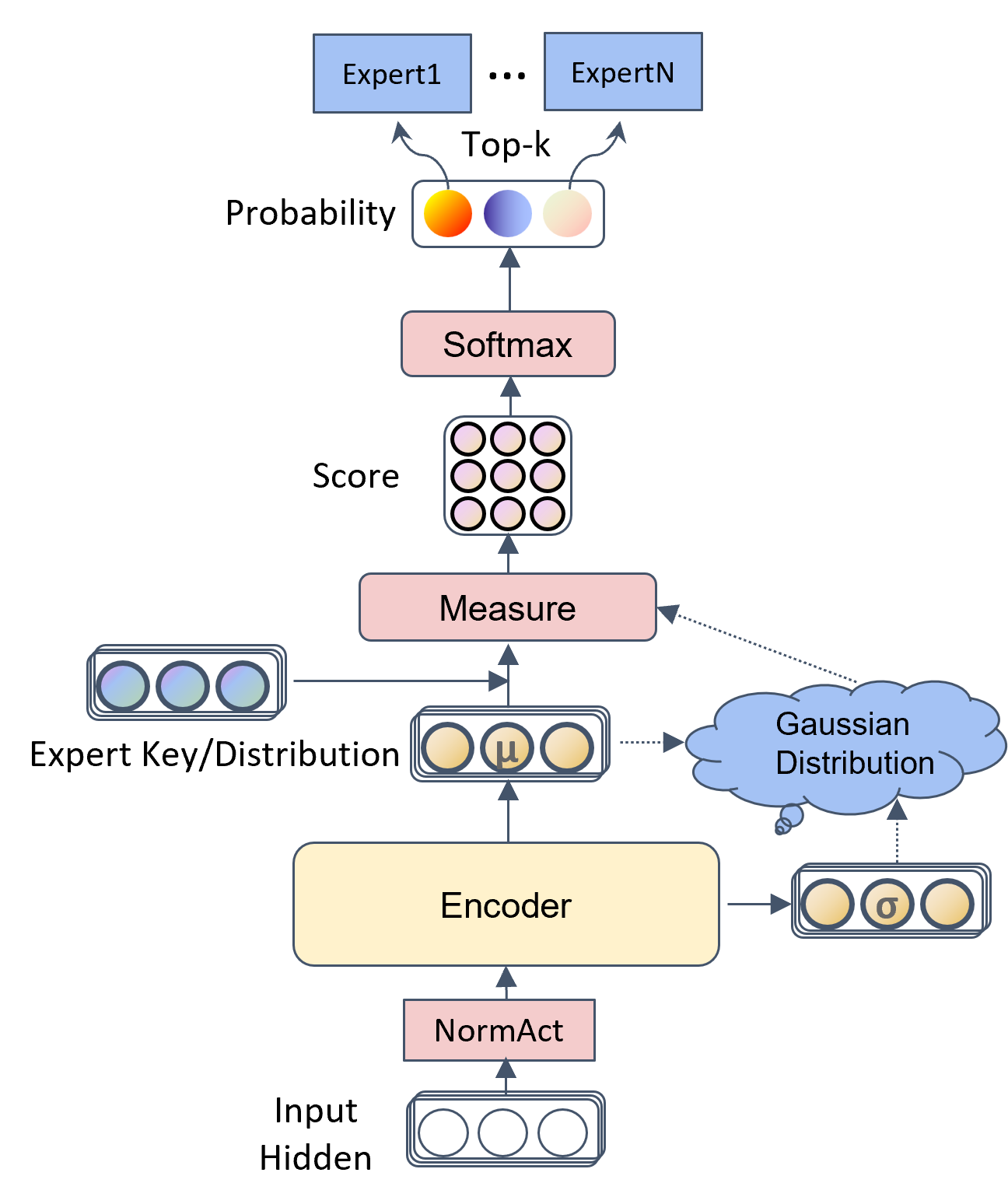}
    \caption{Architecture of Latent Prototype Router}
    \label{fig:routing_arch}
    
\end{figure}

Aiming at solving above questions of vanilla router, we formalize the Router in a more general term. A generalized routing framework consists of three key components (Figure.\ref{fig:routing_arch}):
\begin{align}
    \mathcal{R}(\mathbf{x}) = \mathcal{D}(\mathcal{E}(\mathbf{x}), \mathbf{P})
\end{align}
where $\mathcal{E}$ is Probabilistic/Nonprobabilistic encoder to low dimension latent space, $\mathbf{P}$ is Expert prototypes key, $\mathcal{D}$ is the measurement.
 Our proposed routing framework addresses these limitations through three key innovations:

\paragraph{Non-linear Projection into low-dimension Latent Space.}
To mitigate the curse of dimensionality and improve routing stability, we introduce a simple non-linear encoder that maps token representations into a lower-dimensional latent space:
\begin{equation}
\mathbf{z} = \mathcal{E}(\mathbf{x}) = \mathrm{\text{SiLU}(Norm}(\mathbf{x}))W_1 +\mathbf{b_1}
\end{equation}
where $\mathbf{W}1 \in \mathbb{R}^{d_{\text{model}} \times d_{\text{latent}}}$ with $d_{\text{latent}} \ll d_{\text{model}}$.

To further encourage diversity in the latent space and prevent mode collapse, we introduce a variational encoder that outputs a mean and variance for each token:
\begin{align}
  \boldsymbol{\mu}_t,\,\log\boldsymbol{\sigma}_t^2
    &= \mathcal{E}(\mathbf{h}'_t),\\
  \mathbf{z}_t 
    &= \boldsymbol{\mu}_t + \boldsymbol{\sigma}_t\odot\boldsymbol{\epsilon}_t,
    \quad \boldsymbol{\epsilon}_t\sim\mathcal{N}(\mathbf{0},I).
\end{align}

The encoder is regularized toward a standard Gaussian prior $p(\mathbf{z}) = \mathcal{N}(\mathbf{0}, \mathbf{I})$ by the KL divergence:
\begin{equation}
  \mathcal{L}_{\mathrm{KL}}
  = \frac{1}{T}\sum_{t=1}^T 
    D_{\mathrm{KL}}\bigl(q(\mathbf{z}_t\mid\mathbf{h}'_t)\,\|\,p(\mathbf{z})\bigr),
\end{equation}
where $T$ denotes the number of tokens in a batch and:
\[
  D_{\mathrm{KL}}\bigl(\mathcal{N}(\mu,\sigma^2)\,\|\,\mathcal{N}(0,1)\bigr)
  = \frac{1}{2}\sum_{d=1}^D\bigl(\mu_d^2 + \sigma_d^2 - \log\sigma_d^2 - 1\bigr).
\]

\paragraph{Hyperspherical Initialization of Expert Prototypes.}  
To promote better routing behavior and mitigate expert collapse, we initialize the expert prototype matrix $\mathbf{K} \in \mathbb{R}^{M \times d}$ by sampling each expert key from a standard Gaussian distribution followed by $\ell_2$ normalization:
\[
\mathbf{K}_i = \frac{\mathbf{r}_i}{\|\mathbf{r}_i\|_2}, 
\quad \mathbf{r}_i \sim \mathcal{N}(\mathbf{0}, \mathbf{I}_d).
\]
This effectively places the expert prototypes on the surface of a $d$-dimensional unit hypersphere $\mathbb{S}^{d-1}$. In high-dimensional spaces, this method approximates a uniform distribution over the hypersphere, yielding several beneficial properties on separation and routing dynamic. Expert prototypes uniformly spread on the hypersphere lead to less biased early-stage routing, preventing collapse and ensuring balanced load.
Empirically, we observe that this initialization outperforms both orthogonal and unnormalized random initializations in terms of model performance and load balancing, particularly in the early phases of training.
\paragraph{Diversity Regularization.}
To encourage diverse and well-separated expert prototypes, we impose a diversity regularizer on the encoded tokens matrix $\mathbf{K}\in\mathbb{R}^{B\times d}$, for orthogonality-based diversity regularization:

\begin{equation}
\mathcal{L}_{\text{div}} = \left\| \mathbf{K} \mathbf{K}^\top - \mathbf{I} \right\|_F^2,
\end{equation}

This regularization promotes mutual orthogonality across tokens,  mitigating the collapse of token distribution. We also experiment on diversity loss using euclidean distance and cosine distance.

\paragraph{Alignment Loss via Soft Expert Aggregation.}
To align token latents with their most probable experts, we introduce an explicit reconstruction loss. Given soft routing weights $\mathbf{P}=\mathrm{softmax}(\mathbf{S})$, we compute a weighted sum of projected expert keys.

Given similarity scores $\mathbf{S} \in \mathbb{R}^{B \times M}$ between each token and the expert prototypes, we first compute a soft routing distribution over experts via the softmax function:

\begin{equation}
\mathbf{P} = \text{softmax}(\mathbf{S}),
\end{equation}

where $\mathbf{P}_{ij}$ denotes the probability of token $i$ being routed to expert $j$.

We then compute the softly aggregated expert prototype representation for each token by weighting the projected expert means:

\begin{equation}
\mathbf{K}_{\text{agg}} = \mathbf{P} \cdot\mathbf{K}_{\mu},
\end{equation}

where $\mathbf{K} \in \mathbb{R}^{M \times d}$ are the expert key vectors.

Finally, we define the alignment loss as the $\ell_2$ distance between the detached encoded token latent and the aggregated expert representation:

\begin{equation}
\mathcal{L}_{\text{align}} = \left\| \text{StopGrad}(\mathbf{Z}_{\text{token}}) - \mathbf{K}_{\text{agg}} \right\|_2^2.
\end{equation}

Stopping gradients through the token encoder allows expert prototypes to catch up to the current token cluster assignments without disrupting the learned task features. This further improves specialization by encouraging the prototype space to reflect the underlying token structure. 

\subsubsection{Metric Library}

For the measurement $\mathcal{D}$, we implement the measurement in both geometric distance and distribution distance. For similarity, we have metrics:

\begin{itemize}
    \item \textbf{Cosine}: $\mathcal{S}(\mathbf{z},\mathbf{p}_i) = \frac{\mathbf{z}^T \mathbf{p}_i}{\|\mathbf{z}\|\|\mathbf{p}_i\|}$
    
    \item \textbf{Gaussian Kernel}:
    $\mathcal{S}(\mathbf{z},\mathbf{p}_i) = \exp \left( -\frac{\|\mathbf{z} - \mathbf{p}_i\|^2}{2\sigma^2} \right)$
    \item \textbf{Multi-Head Dot Product Attention.}Given token queries $\mathbf{Q}$ and expert keys $\mathbf{K}$, the attention scores for head $h$ are:

\begin{equation}
\mathrm{Attn}^{(h)}(\mathbf{Q}, \mathbf{K}) = \frac{\mathbf{Q}^{(h)} \cdot \mathbf{K}^{(h)\top}}{\sqrt{d_h}},
\end{equation}

where $d_h$ is the dimensionality per head. The final similarity is averaged across heads:

\begin{equation}
\mathcal{S}(\mathbf{q}, \mathbf{K}) = \frac{1}{H} \sum_{h=1}^{H} \mathrm{Attn}^{(h)}.
\end{equation}

\end{itemize}

For distribution distance, we assume expert key also follow gaussian distributions. Therefore:
\begin{itemize}
\item The 2-Wasserstein distance between diagonal Gaussians is:
\begin{equation}
\mathrm{Dist} = \mathcal{W}_2^2(\mathcal{N}_1, \mathcal{N}_2) = \| \boldsymbol{\mu}_1 - \boldsymbol{\mu}_2 \|_2^2 + \| \sqrt{\boldsymbol{\sigma}_1^2} - \sqrt{\boldsymbol{\sigma}_2^2} \|_2^2.
\end{equation}
\item  For two diagonal Gaussian distributions $\mathcal{N}(\boldsymbol{\mu}_1, \boldsymbol{\sigma}_1^2)$ and $\mathcal{N}(\boldsymbol{\mu}_2, \boldsymbol{\sigma}_2^2)$, the KL divergence has a closed-form expression:
\begin{equation}
\mathrm{Dist} = \mathrm{KL}\left( \mathcal{N}_1 \,\|\, \mathcal{N}_2 \right) = \frac{1}{2} \sum_{i=1}^d \left[ \log\left( \frac{\sigma_{2,i}^2}{\sigma_{1,i}^2} \right) + \frac{\sigma_{1,i}^2 + (\mu_{1,i} - \mu_{2,i})^2}{\sigma_{2,i}^2} - 1 \right].
\end{equation}
\item  For JS-Divergence, we have:
\begin{equation}
    \mathrm{Dist} = \text{JS}(\left( \mathcal{N}_1 \,\|\, \mathcal{N}_2 \right) = \frac{1}{4} \left( \ln\frac{(\sigma_1^2 + \sigma_2^2)^2}{4\sigma_1^2\sigma_2^2} + \frac{\sigma_1^2 + (\mu_1 - \mu_0)^2}{\sigma_0^2} + \frac{\sigma_2^2 + (\mu_2 - \mu_0)^2}{\sigma_0^2} - 2 \right)
\end{equation}
\item  For Hellinger distance, we have:
\begin{equation}
   \mathrm{Dist} =  H( \mathcal{N}_1 , \mathcal{N}_2) = \frac{1}{\sqrt{2}} \sqrt{1 - \frac{2\sigma_1\sigma_2}{\sigma_1^2 + \sigma_2^2} \exp\left( -\frac{1}{4} \frac{(\mu_1 - \mu_2)^2}{\sigma_1^2 + \sigma_2^2} \right)}
\end{equation}
 \end{itemize}

\subsection{Total loss}
The complete optimization objective combines:
\begin{align}
\mathcal{L} &= \mathcal{L}_{\text{task}} + \beta_{rs}( \beta_1 \mathcal{L}_{\text{div}} + \beta_2 \mathcal{L}_{\text{align}} +  \beta_3 \mathcal{L}_{\text{KL}}),
\end{align} where $\beta_{rs}$ control the strength of  all regulation. With these regulation, we obtain a MoE model with much better load balance and better task performance, which we will show in the experiment section.

\section{Experiment}
In this section, we evaluate the proposed Latent Prototype Routing (LPR) mechanism on several mainstream large language model (LLM) architectures. We first describe the experimental setup and hyperparameter choices, then report quantitative results, followed by an analysis of routing quality, expert utilization, and computational efficiency.

\subsection{Experimental Setup}
\paragraph{Datasets.} We conduct experiments on the fineweb\citep{fineweb} sample-100BT subset. Specifically, we use 1 billion tokens for training mainstream MoE models (Table\ref{tab:mainstream-results}), and 100 million tokens for all other ablations. Scaling the training data and model size is generally expected to improve absolute performance. We believe that the benefits of LPR would become even more pronounced in large-scale regimes, as balanced routing yields more efficient parameter utilization.

\paragraph{Model Architecture.}
Our experiments use MoE-augmented Transformers ranging from 0.1B to 0.6B parameters. We evaluate against three strong open-source baselines: DeepSeek-V3\cite{deepseekai2025deepseekv3technicalreport}, Mixtral\cite{jiang2024mixtralexperts}, and Qwen3MoE\cite{yang2025qwen3technicalreport}. Our proposed model, Qwen3MoE-LPR, is a modified version of Qwen3MoE incorporating the latent prototype router. Ablation are based on Qwen3MoE-LPR model. Settings and detailed architecture hyperparameters are provided in Appendix~A.

\paragraph{Training and Hyperparameters.}

We use a warmup–stable–decay learning rate scheduler, with 5\% of total steps dedicated to linear warmup, 25\% to cosine decay, and the remaining steps using a stable learning rate. LPR-specific hyperparameters are set as follows: latent dimension $d_{\text{latent}}=16$, KL weight $\beta_{KL}=0.01$, alignment loss weight $\beta_{align}=0.05$, diversity regularization weight $\beta_{div}=1.0$, and a global regularization scale $\beta_{rs}=0.01$. All models are trained with AdamW ($\beta_1=0.9$, $\beta_2=0.95$), weight decay $0.1$, and gradient norm clipping at $1.0$. Please refer to Appendix~A for the full list of training hyperparameters.

\paragraph{Metrics}
We use several metrics to evaluate load balance and routing quality across experts:
\begin{itemize}
\item \textbf{Gini Coefficient:}
\begin{equation}
   \quad
\mathrm{Gini} = \frac{1}{n \sum_{i=1}^n l_i} \sum_{i=1}^n (2i - n - 1) \, l_{(i)}
\end{equation}
where \(l_{(i)}\) denotes the $i$-th smallest expert load and $n$ is the number of experts. A Gini coefficient of $0$ indicates perfect balance, while $1$ denotes extreme imbalance.
\item \textbf{Min-Max Ratio:}
\begin{equation}
    \quad
\mathrm{MinMaxRatio} = \frac{\min_i l_i}{\max_i l_i + \epsilon}
\end{equation}
where $\epsilon$ is a small constant to prevent division by zero. Values close to $1$ imply uniform load across experts; values approaching $0$ indicate that some experts receive vastly fewer tokens.
\end{itemize}

\subsection{Balanced Routing via Latent Prototype Routing}
\label{sec:mainstream_results}

Table~\ref{tab:mainstream-results} compares routing methods on the C4 validation set, evaluating \textbf{test loss}, expert load balance (GINI coefficient), and expert utilization (Min--Max ratio). Baseline implementations follow standard practices: Qwen3Moe and Mixtral use auxiliary balance losses, while DeepSeekV3 employs bias correction\cite{wang2024auxiliarylossfreeloadbalancingstrategy}. Our LPR method uses only its intrinsic regularization.

\paragraph{Load Imbalance in Standard MoEs} 
Baseline models exhibit severe routing imbalance despite competitive performance. DeepSeekV3-0.6B achieves test loss 3.673 but suffers high imbalance (GINI=0.790, Min--Max=$6.41 \times 10^{-9}$). Similarly, Qwen3Moe-0.6B (test loss=3.666) shows GINI=0.707 and near-zero Min--Max ratio ($1.27 \times 10^{-16}$), indicating chronic expert underutilization.

\paragraph{LPR Achieves Near-Ideal Balancing}
Our Latent Prototype Router (LPR) dramatically improves load distribution while maintaining competitive performance:
\begin{itemize}
    \item \textbf{DeepSeekMoe-LPR}: Reduces GINI by 95\% (0.790 $\rightarrow$ 0.036) and improves Min--Max by 8 orders of magnitude ($6.41 \times 10^{-9} \rightarrow 0.724$) with minimal loss increase (3.720 vs 3.673)
    \item \textbf{Qwen3Moe-LPR w/ Hyperspherical}: Maintains near-identical performance (3.685 vs 3.666) while achieving balanced utilization (Min--Max=0.597, GINI=0.057)
    \item \textbf{Mixtral-LPR}: Improves Min--Max ratio by 5 orders of magnitude (3.33e-6 $\rightarrow$ 0.649) and reduces GINI by 93\% (0.635 $\rightarrow$ 0.047)
\end{itemize}

\paragraph{Hyperspherical Initialization} Hyperspherical initialization provides consistent gains: Qwen3Moe-LPR with hyperspherical init outperforms its counterpart (3.685 vs 3.697 test loss) under identical balancing.

\begin{table}[t]
\centering
\caption{\textbf{Routing method comparison on C4 validation set.} LPR achieves near-perfect load balance (GINI $\approx$ 0.05). Min--Max ratios improve by 5--8 orders of magnitude versus baselines.}
\label{tab:mainstream-results}
\small
\begin{tabular}{@{}lcccc@{}}
\toprule
\textbf{Method} & \textbf{Test Loss} $\downarrow$ & \textbf{GINI} $\downarrow$ & \textbf{Min-Max} $\uparrow$ \\
\midrule
Mixtral-0.6B (128-8) & \textbf{3.683} & 0.635 & $3.33 \times 10^{-6}$ \\
Mixtral-LPR-0.6B (w/o init) & 3.747 & \textbf{0.047} & \textbf{0.649} \\
\midrule
DeepSeekV3-0.6B (128-8) & \textbf{3.673} & 0.790 & $6.41 \times 10^{-9}$ \\
DeepSeekMoe-LPR (w/o init) & 3.720 & \textbf{0.036} & \textbf{0.724} \\
\midrule
Qwen3Moe-0.6B (128-8) & \textbf{3.666} & 0.707 & $1.27 \times 10^{-16}$ \\
Qwen3Moe-LPR (w/ init) & 3.685 & 0.057 & 0.597\\
Qwen3Moe-LPR (w/o init) & 3.697 & \textbf{0.039} & \textbf{0.696} \\
\bottomrule
\end{tabular}
\end{table}

\paragraph{Performance of a Balanced Routing} While LPR achieves near-ideal load balancing (see Table~\ref{tab:mainstream-results}), the marginal performance improvement observed at current model and training scales reflects a fundamental antagonistic trade-off:

\begin{itemize}
\item On one hand, balanced routing increases the effective number of parameters actively participating in training, thereby improving long-term model capacity and potential performance.
\item On the other hand, excessive balancing conflicts with the inherently uneven and clustered nature of natural language token distributions, which can degrade expert specialization and harm model effectiveness.
\end{itemize}

This antagonism means that although balanced routing methods like LPR maximize expert utilization and scaling potential, over-balancing can inadvertently reduce specialization benefits and cause performance to plateau or even decline. Therefore, careful design of routing mechanisms is crucial to strike an optimal balance between load uniformity and expert specialization (see Appendix~\ref{app:scaling_advantage} for conceptual illustrations).

Nevertheless, it is important to emphasize that balanced MoE architectures enable significantly more efficient hardware utilization, which in turn can unlock the potential to achieve comparable performance with fewer parameters and lower computational cost.

\paragraph{Conclusion} LPR fundamentally solves MoE routing imbalance without compromising model quality. The 10--20$\times$ GINI reduction and 5--8 order-of-magnitude Min--Max improvements demonstrate superior hardware utilization and training stability – critical for scaling sparse models.

\subsection{Ablation Studies}
To quantify the contribution of each component in our Latent Prototype Router (LPR) and to understand the sensitivity of key hyperparameters, we perform a comprehensive suite of ablation experiments on the C4 validation set. We report test loss, load imbalance measured by GINI coefficient, and min–max expert usage ratio.  

\paragraph{Component-wise Ablation}
Table~\ref{tab:ablation} summarizes the impact of removing individual modules from the full LPR model.  
\begin{itemize}
  \item \textbf{KL Regularization} (\emph{w/o KL}, $\beta=0$): disabling the variational KL term dramatically degrades balance (GINI ↑ to 0.115, min–max ↓ to 0.304) and lowers test loss to 4.82, indicating that without this regularizer the router overfits to high-density clusters, leading to skewed dispatch.  
  \item \textbf{Alignment Loss} (\emph{w/o Align Loss}): removing the token–prototype alignment objective yields similar imbalance as the KL-only ablation (GINI 0.115, min–max 0.286), confirming that explicit alignment is central to both performance and balance.  
  \item \textbf{Diversity Loss} (\emph{w/o Diversity Loss}): omitting the orthogonality-based diversity penalty drastically increases imbalance (GINI 0.716) and causes the worst test loss (5.01), demonstrating that enforcing prototype separation is essential to prevent expert collapse and hence increase the model capacity.  
\end{itemize}
\begin{table}[ht]
\centering
\caption{Ablation study on the Latent Prototype Router (LPR) components. 
Each variant removes a specific component from the full LPR model to assess its contribution.
Results are reported on the C4 validation set in terms of test loss and expert load imbalance, measured by the GINI coefficient and Min-Max ratio.
}
\label{tab:ablation}
\begin{tabular}{lccc}
\toprule
\textbf{Variant} & \textbf{Test loss} $\downarrow$& \textbf{GINI} $\downarrow$ & \textbf{Min-Max} $\uparrow$\\
\midrule
Full LPR                     & 4.86 & \textbf{0.06} & \textbf{0.595}\\
\quad w/o KL ($\beta=0$)     & \textbf{4.82} & 0.115 &0.304 \\
\quad w/o Align Loss         & 4.83 & 0.115 & 0.286 \\
\quad w/o Diversity Loss     & 5.01 & 0.716& 0.002 \\
\bottomrule
\end{tabular}
\end{table}

\paragraph{Latent Dimension and Regularization Strength}

Next, we explore how the encoder’s latent dimensionality and the load-balancing regularization weight affect performance. Table.\ref{latend} shows that a latent dimension of 16 achieves the best trade-off (test loss 4.869, GINI 0.060, min–max 0.595), while both smaller and larger dimensions underperform or introduce imbalance. On regularization strength (Table.\ref{regula}), we find $\beta=0.01$ to be optimal; weaker strength (0.00) fails to balance loads, and excessively strong regularization over-constrains clustering, leading to higher test loss and either over- or under-regularization artifacts.  

\begin{table}[t]
\centering
\label{tab:triangular-layout}
\begin{minipage}{0.48\textwidth}
\centering
\captionof{table}{Effect of encoder's latent dimension.}
\begin{tabular}{lccc}
\toprule
\textbf{Latent Dim} & \textbf{Test Loss} $\downarrow$& \textbf{GINI} $\downarrow$ & \textbf{Min-Max} $\uparrow$\\
\midrule
4   & 5.085& 0.122 & 0.385 \\
8   & 4.927& 0.085 & 0.480 \\
16  & 4.869& \textbf{0.060} & \textbf{0.595} \\
32  & \textbf{4.828}& 0.070  & 0.5247    \\
64  & 4.874& 0.063 & 0.525 \\
128 & 4.891& 0.074 & 0.507 \\
256 & 4.902& 0.093  & 0.395 \\
\bottomrule
\label{latend}
\end{tabular}
\end{minipage}
\hfill
\begin{minipage}{0.48\textwidth}
\centering
\captionof{table}{Effect of regularization strength. $\dagger$: Not yet saturated at the end of training, suggesting more training time needed}
\begin{tabular}{lccc}
\toprule
\textbf{Reg. Strength} & \textbf{Test Loss} $\downarrow$& \textbf{GINI} $\downarrow$ & \textbf{Min-Max} $\uparrow$\\
\midrule
0.00 & 4.995& 0.72     & 0.0009     \\
0.01 &  \textbf{4.870}& 0.060 & 0.595 \\
0.04 & 5.060& \textbf{0.043} & \textbf{0.668} \\
0.10 & 5.234& 0.044 & 0.662 \\
0.50 $\dagger$ & 5.752& 0.05     & 0.628     \\
\bottomrule
\label{regula}
\end{tabular}
\end{minipage}
\end{table}

\paragraph{Flexibility and Diversity Measures}
In Table.\ref{scality}, we further validate that our router maintains low variance and competitive GINI ratios across extreme expert counts (up to 512 experts with $k=8$), whereas a no-reg variant catastrophically collapses (load variance $6.19\times10^4$, GINI 0.985). Results show our method provide a more flexible design space for MoE architecture. Table.\ref{div_me} (lower right) compares different diversity measures: orthogonality delivers the best balance/performance trade-off (test loss 4.86, GINI 0.060), cosine and Euclidean each bias towards either balance or specialization.

\begin{table}[t]
  \begin{minipage}{0.45\textwidth}
    \centering
    \captionof{table}{Effect of number of experts. Experiment show that our algorithm could suppress expert load imbalance in large extend even in extreme conditions.}
    \begin{tabular}{lccc}
      \toprule
      \textbf{N-k Setting}  & \textbf{GINI} $\downarrow$ & \textbf{Min-Max} $\uparrow$\\
      \midrule
      128-8         &\textbf{ 0.099}    & \textbf{0.412}       \\
      256-8     & 0.155   & 0.245   \\
      512-8        & 0.249   & 0.059       \\
      512-4        & 0.347    & 0.018       \\
      512-1       & 0.322   & 0.047 \\
      512-1-no reg.  &  0.9853   & 9.3e-22      \\
      \bottomrule
      \label{scality}
    \end{tabular}
  \end{minipage}
  \hspace{1cm}
  \begin{minipage}{0.45\textwidth}
    \centering
    \captionof{table}{Effect of diversity measure.}
    \begin{tabular}{lccc}
      \toprule
      \textbf{Diversity Mea.} & \textbf{Test Loss} $\downarrow$ & \textbf{GINI} $\downarrow$ & \textbf{Min-Max} $\uparrow$\\
      \midrule
      Cosine       & 5.11 &0.482   & 0.037   \\
      Orthogonal   & \textbf{4.86}   & \textbf{0.06}   & \textbf{0.595}      \\
      Euclidean    & 6.745   & 0.263  & 0.111   \\
      \bottomrule
      \label{div_me}
    \end{tabular}
  \end{minipage}
\end{table}

\paragraph{Similarity/Divergence in Routing}
In Table~\ref{tab:metric-table}, we benchmark geometric (left) versus distributional (right) similarity metrics for routing. Cosine similarity remains a solid default (test loss 4.855, GINI 0.082), but our orthogonality-based kernel outperforms more expensive measures. Among distributional divergences, KL divergence offers reasonable PPL (4.881) and balance (GINI 0.261).

\begin{table}[ht]
  \centering
  \caption{
  Comparison of different similarity/distance measures used in expert routing.
  Left: geometric-based measures. Right: distribution-based measures.
  }
  \label{tab:metric-table}
  \begin{tabular*}{\textwidth}{@{\extracolsep{\fill}}cc}
    \begin{tabular}{lccc}
      \toprule
      \textbf{Measure} & \textbf{Test Loss} $\downarrow$& \textbf{GINI} $\downarrow$ & \textbf{Min-Max} $\uparrow$\\
      \midrule
      Cosine           & \textbf{4.855} & \textbf{0.082}  & \textbf{0.595} \\
      Gaussian Kernel  & 4.908 & 0.269  & 0.139 \\
      Mahalanobis      & 4.910 & 0.246  & 0.111 \\
      Cross-Attention  & 4.878 & 0.574  & 0.007 \\
      \bottomrule
    \end{tabular}
    \begin{tabular}{lccc}
      \toprule
      \textbf{Measure} & \textbf{Test Loss} $\downarrow$& \textbf{GINI} $\downarrow$ & \textbf{Min-Max} $\uparrow$\\
      \midrule
      Wasserstein      & 4.884& 0.29 & 0.067 \\
      Hellinger        & 4.964& 0.364  & 0.043 \\
      JS Divergence    & 4.979& 0.298  & 0.08 \\
      KL Divergence    & \textbf{4.881} & \textbf{0.261}  & \textbf{0.098} \\
      \bottomrule
    \end{tabular}
  \end{tabular*}
\end{table}

\textbf{Summary.} Our ablations confirm that each design choice—from hyperspherical initialization to orthogonality regularization—plays a critical role in balancing specialization against load uniformity. The latent dimension and regularization weight must be jointly tuned to achieve optimal language modeling performance without sacrificing resource utilization.
\newpage

\section{Related Works}
\subsection{Transformers and the Emergence of Mixture-of-Experts}
The transformer architecture introduced by Vaswani et al. \cite{vaswani2017attention} has become the backbone of modern large language models (LLMs). By leveraging self-attention mechanisms, transformers enable parallel processing of sequence data, fundamentally advancing natural language understanding and generation tasks. However, dense transformer models face inherent limitations: quadratic computational complexity with respect to sequence length \cite{vaswani2017attention}, exorbitant memory requirements for models with billions of parameters \cite{lepikhin2020gshard}, diminishing performance returns as model scale increases \cite{hoffmann2022training}, and high deployment costs \cite{fedus2021switch}.
Mixture-of-Experts (MoE) has emerged as an effective solution to address these challenges. Originating from the work of Jacobs et al. \cite{jacobs1991adaptive}, MoE models dynamically route inputs to a subset of specialized "expert" networks, enabling exponential parameter scaling with sublinear computational cost. A pivotal advancement came with the integration of MoE into transformer architectures, replacing traditional feed-forward networks (FFNs) with MoE layers \cite{lepikhin2020gshard}. Lepikhin et al. \cite{lepikhin2020gshard} demonstrated the feasibility of this approach with GShard, a 600B-parameter multilingual transformer with MoE that outperformed dense models in terms of efficiency. Fedus et al. \cite{fedus2021switch} further pushed the envelope with Switch Transformer, scaling to trillion-parameter models using top-1 routing.
\subsection{Recent Advances in MoE Architectures for LLMs}
Over the past five years, research on MoE architectures for LLMs has seen significant diversification. Mixture of Latent Experts (MoLE) \cite{liu2025molaemixturelatentexperts} factorizes experts into a shared latent space, reducing model parameters while preserving representational capacity. Cluster-driven Expert Pruning (C-Prune) \cite{guo2025c} addresses intra-layer homogeneity and inter-layer similarity to enable task-specific model compression. Symbolic-MoE \cite{chen2025symbolicmixtureofexpertsadaptiveskillbased} introduces a gradient-free framework for instance-level expert mixing based on text-based symbolic reasoning. AdaMoE \cite{zeng2024adamoe} proposes token-adaptive routing with "null experts," reducing compute flops by 14.5\% while improving accuracy on the ARC-C dataset.
\subsection{Researches on Expert Load Balancing}
Most existing work on expert load balancing in MoE models has focused on two main directions: designing auxiliary balancing losses and pruning underutilized experts. For example, auxiliary loss terms that encourage more uniform routing across experts have been widely adopted. In contrast, DeepSeek\cite{wang2024auxiliarylossfreeloadbalancingstrategy} proposes an auxiliary-free strategy to achieve balanced expert utilization. However, research on the routing mechanism itself remains limited. One reason is that the vanilla top-$k$ gating mechanism is overly simplistic and lacks a unified design framework. This restricts further exploration, as it is often unclear where and how to intervene in the router for substantial improvements. Our work aims to address this gap by formalizing a general routing framework and introducing explicit constraints that can systematically improve routing behavior.
\section{Conclusion}
In this paper, we introduced a simple yet effective method for addressing the load-balancing problem, a critical challenge that has long hindered the efficient utilization of parameters and inference in Mixture-of-Experts (MoE) models. We identified that the root causes of this issue lie in the excessive dimensionality of the routing space, the simplistic nature of linear encoders, and the lack of explicit modeling of token clusters and their distribution. By tackling these questions one by one, our approach significantly mitigates imbalances in expert usage. Moreover, we observed that enhancing token separation improves both load balance and expert specialization, thereby boosting overall model performance. However, as separation becomes too aggressive, gains in performance come at the cost of load balance, suggesting that token distribution is naturally non-uniform and should be accounted for when designing future MoE routing strategies and hardware implementation.
\bibliographystyle{unsrt}  
\bibliography{references}  
\newpage

\appendix

\section{Model setting}

We compare three Mixture-of-Experts (MoE) architectures at the 0.6B parameter scale: Qwen3MoE, DeepseekV3, and Mixtral, along with their corresponding variants enhanced with the proposed Latent Prototype Router (LPR). All models share consistent training hyperparameters, including the number of training tokens (100M), maximum sequence length (1024), and optimizer settings (Adam with $\beta_1=0.9$, $\beta_2=0.95$). 

As shown in Table~\ref{tab:config-comparison}, the LPR-augmented variants introduce additional routing-specific components such as a latent projection head (`router\_latent\_dim`), orthogonality-based diversity regularization, KL regularization, and an optional alignment loss. In these variants, the EMA update of expert key is disabled, and replaced with a direct latent alignment scheme using a unit-ball constraint and cosine-based similarity metrics.

For all variants, the number of experts is fixed at 128, and 8 experts are selected per token. The LPR configurations aim to enhance expert specialization while promoting load balance through latent-space regularization. Architecture-specific differences (e.g., hidden size, attention head configuration) are preserved from their respective baselines for a fair comparison. Unless otherwise specified, all ablation studies are conducted using the Qwen3MoE-LPR configuration from Table.\ref{tab:config-comparison}, with only the relevant hyperparameter(s) varied.

\begin{table}[ht]
\caption{
Model configuration comparison across different 0.6B parameter-scale architectures, including Qwen3MoE, DeepseekV3, and Mixtral, with and without Latent Prototype Router (LPR). ``--'' denotes that the feature is not used in the respective variant.
}
\centering
\scriptsize
\resizebox{\linewidth}{!}{
\begin{tabular}{lcccccc}
\toprule
\textbf{Configuration} & \textbf{Qwen3MoE} & \textbf{Qwen3MoE-LPR} & \textbf{DeepseekV3} & \textbf{DeepseekV3-LPR} & \textbf{Mixtral} & \textbf{Mixtral-LPR} \\
\midrule
\textbf{\#Params} & 0.6B & 0.6B & 0.6B & 0.6B & 0.6B & 0.6B \\
Vocab Size & 151936 & 151936 & 151936 & 151936 & 151936 & 151936 \\
Hidden Size & 1204 & 1204 & 1024 & 1024 & 1024 & 1024 \\
Head Dim & 128 & 128 & -- & -- & 64 & 64 \\
QK NoPE Head Dim & -- & -- & 64 & 64 & -- & -- \\
QK RoPE Head Dim & -- & -- & 128 & 128 & -- & -- \\
\#Attention Heads & 16 & 16 & 8 & 8 & 16 & 16 \\
\#Key/Value Heads & 4 & 4 & 8 & 8 & 4 & 4 \\
Hidden Activation & SiLU & SiLU & SiLU & SiLU & SiLU & SiLU \\
MoE Intermediate Size & 128 & 128 & 128 & 128 & 128 & 128 \\
\#Shared Experts & -- & -- & 5 & 5 & -- & -- \\
Experts per Top-k & 8 & 8 & 8 & 8 & 8 & 8 \\
Total Experts & 128 & 128 & 128 & 128 & 128 & 128 \\
Aux Loss Coef & 1e-3 & 0 & -- & 0 & 1e-3 & 0 \\
LPR Loss Coef & 0 & 1e-2 & -- & 0.01 & 0 & 0.01 \\
LPR Latent Dim & -- & 16 & -- & 16 & -- & 16 \\
EMA Update & -- & False & -- & -- & -- & -- \\
EMA Decay & -- & 0.9 & -- & 0.9 & -- & 0.9 \\
Similarity Metric & -- & VectorSim & -- & VectorSim & -- & VectorSim \\
Diversity Type & -- & Orthogonal & -- & Orthogonal & -- & Orthogonal \\
Diversity Loss Weight & -- & 1 & -- & 1 & -- & 1 \\
KL Loss Weight & -- & 0.01 & -- & 0.01 & -- & 0.01 \\
Align Loss Weight & -- & 0.1 & -- & 0.1 & -- & 0.1 \\
Total LPR Loss Coef & -- & 0.01 & -- & 0.01 & -- & 0.01 \\
Unit Ball Constraint & -- & True & -- & True & -- & True \\
Adam $\beta_1$ & 0.9 & 0.9 & 0.9 & 0.9 & 0.9 & 0.9 \\
Adam $\beta_2$ & 0.95 & 0.95 & 0.95 & 0.95 & 0.95 & 0.95 \\
Max Length & 1024 & 1024 & 1024 & 1024 & 1024 & 1024 \\
Training Tokens & 100M & 100M & 100M & 100M & 100M & 100M \\
Init Learning Rate & 0.001 & 0.001 & 0.001 & 0.001 & 0.001 & 0.001 \\
Min LR Ratio & 0.05 & 0.05 & 0.05 & 0.05 & 0.05 & 0.05 \\
Warmup Range & 5\% & 5\% & 5\% & 5\% & 5\% & 5\% \\
Stable Phase & 75\% & 75\% & 75\% & 75\% & 75\% & 75\% \\
\bottomrule
\end{tabular}
}
\label{tab:config-comparison}
\end{table}

\newpage
\section{Scaling up and the Conflicts between Expert Specialization and Load Balance}
\label{app:scaling_advantage}

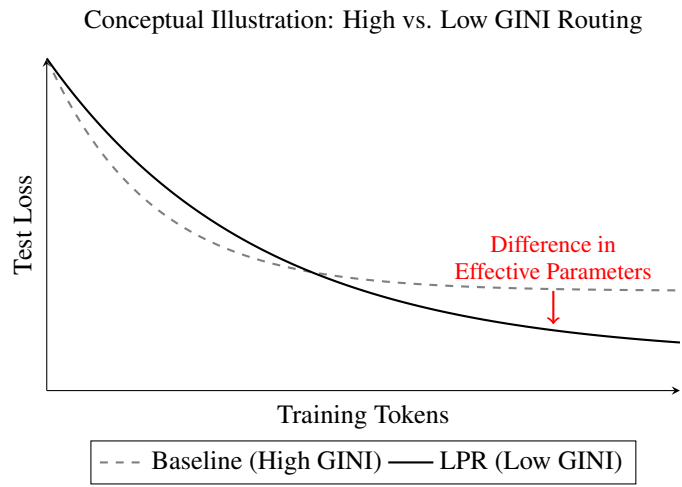
\begin{figure}[htbp]
  \centering
  \begin{tikzpicture}
    \begin{axis}[
        axis lines = left,
        xlabel = {Training Tokens},
        ylabel = {Test Loss},
        xtick=\empty,
        ytick=\empty,
        xmin=0, xmax=1,
        ymin=0.0, ymax=1.0,
        width=10cm,
        height=6cm,
        legend style={at={(0.5,-0.15)}, anchor=north, legend columns=2},
        title = {Conceptual Illustration: High vs. Low GINI Routing},
        every axis plot/.append style={thick}
    ]

    \addplot[dashed, gray] 
        plot[domain=0:1, samples=200] 
        {1 - 0.7 * (1 - exp(-6 * x))};
    \addlegendentry{Baseline (High GINI)}

    \addplot[black] 
        plot[domain=0:1, samples=200] 
        {1 - 0.9 * (1 - exp(-3 * x))};
    \addlegendentry{LPR (Low GINI)}

    \draw[->, thick, red] 
        (axis cs:0.8,0.3) -- (axis cs:0.8,0.2);
    \node[red, align=center, font=\small] at (axis cs:0.8,0.4) 
        {Difference in\\Effective Parameters};

    \end{axis}
  \end{tikzpicture}
  \caption{Conceptual illustration of the impact of expert utilization on convergence. Although high-GINI models converge faster, LPR (low GINI) achieves better final performance due to more effective expert participation.}
  \label{fig:gini_concept}
\end{figure}

Figure~\ref{fig:gini_concept} conceptually illustrates the relationship between training scale and routing mechanism effectiveness. The horizontal axis corresponds to the total number of training tokens, representing training scale, while the vertical axis measures model performance (e.g., test loss or accuracy).

When parameter utilization efficiency is fixed, increasing the total parameter count exacerbates inefficiency and parameter waste. Consequently, high-GINI models show substantially worse performance compared to low-GINI models as model scale grows, since fewer experts contribute effectively despite the larger overall capacity.

This highlights the importance of routing strategies that promote broad and balanced expert participation. Particularly in large-scale training, maximizing the effective use of parameters is critical to achieve improved convergence and final model quality.

\vspace{1em}

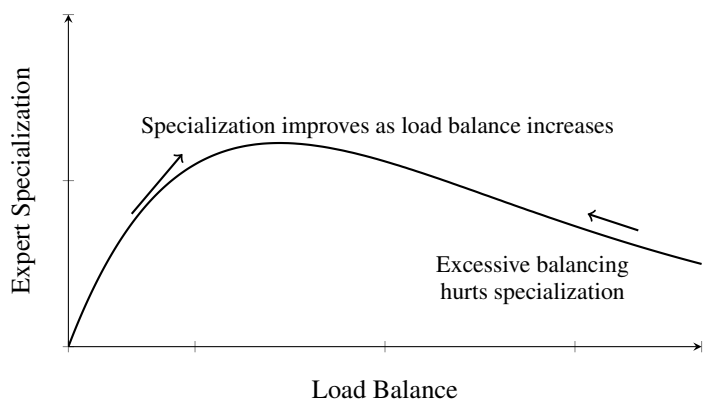
\begin{figure}[htbp]
  \centering
  \begin{tikzpicture}
    \begin{axis}[
        width=10cm,
        height=6cm,
        xlabel={Load Balance},
        ylabel={Expert Specialization},
        xmin=0, xmax=1,
        ymin=0, ymax=1,
        xtick={0,0.2,0.5,0.8,1},
        ytick={0,0.5,1},
        xticklabels=\empty,  
        yticklabels=\empty,  
        axis lines=left,
        title={Figure 3: The Trade-off between Expert Specialization and Load Balance},
        every axis plot/.append style={thick, smooth},
        ]
      \addplot[
        black,
        domain=0:1,
        samples=200,
      ] {5*x*exp(-3*x)};
      
      \node[anchor=south west, font=\small] at (axis cs:0.1,0.6) {Specialization improves as load balance increases};
      \draw[->, thick] (axis cs:0.1,0.4) -- (axis cs:0.18,0.58);
      
      \node[anchor=north east, font=\small, align=center] at (axis cs:0.9,0.3) {Excessive balancing \\ hurts specialization};
      \draw[->, thick] (axis cs:0.9,0.35) -- (axis cs:0.82,0.4);
    \end{axis}
  \end{tikzpicture}
  \caption{The trade-off between expert specialization and load balance. Initially, improving load balance enhances expert specialization as tokens become more separable when redistributed from dense clusters. However, beyond a certain point, forcing balance reduces specialization due to the inherently uneven nature of token distributions.}
  \label{fig:expert_specialization}
\end{figure}

Figure~\ref{fig:expert_specialization} illustrates the inherent trade-off between expert specialization and load balance. At low levels of load balancing, an increasing balance generally suggest that tokens are better separated from dense distributions, which promotes higher expert specialization as well as load balance.

However, as load balance becomes increasingly enforced, the natural unevenness of token distributions makes it difficult to maintain specialization. Excessive balancing forces experts to process tokens that do not align well with their specialization, leading to a decline in expert effectiveness.

Together, Figures~\ref{fig:gini_concept} and \ref{fig:expert_specialization} demonstrate that while balanced expert utilization is crucial for large-scale model performance, it must be carefully managed to preserve expert specialization, as over-balancing can degrade model effectiveness. Future work should explore routing mechanisms that dynamically balance this trade-off to maximize both specialization and load balance.


\end{document}